\documentclass[10pt,twocolumn,letterpaper]{article}

\usepackage{iccv}
\usepackage{times}
\usepackage{epsfig}
\usepackage{graphicx}
\usepackage{amsmath}
\usepackage{amssymb}
\usepackage{subfigure}
\usepackage{diagbox}
\usepackage{overpic}
\usepackage{adjustbox}
\usepackage{breqn}
\usepackage{multirow}
\usepackage{multicol}
\usepackage{setspace}

\usepackage{enumitem}

\usepackage[pagebackref=true,breaklinks=true,letterpaper=true,colorlinks,bookmarks=false]{hyperref}

\iccvfinalcopy 
\def\iccvPaperID{0000} 

\ificcvfinal\fi
\begin{document}

\title{FDLite: A Single Stage Lightweight Face Detector Network}

\author{Yogesh Aggarwal\\
Indian Institute of Technology Guwahati\\
Assam India\\
{\tt\small yogesh\_aggarwal@iitg.ac.in}
\and
Prithwijit Guha \\
Indian Institute of Technology Guwahati\\
Assam India\\
{\tt\small pguha@iitg.ac.in}
}

\maketitle

\begin{abstract}
Face detection is frequently attempted by using heavy pre-trained backbone networks like ResNet-50/101/152 and VGG16/19. Few recent works have also proposed lightweight detectors with customized backbones, novel loss functions and efficient training strategies. The novelty of this work lies in the design of a lightweight detector while training with only the commonly used loss functions and learning strategies. The proposed face detector grossly follows the established RetinaFace architecture. The first contribution of this work is the design of a customized lightweight backbone network (BLite) having 0.167M parameters with 0.52 GFLOPs. The second contribution is the use of two independent multi-task losses. The proposed lightweight face detector (FDLite) has 0.26M parameters with 0.94 GFLOPs. The network is trained on the WIDER FACE dataset. FDLite is observed to achieve 92.3\%, 89.8\%, and 82.2\% Average Precision (AP) on the easy, medium, and hard subsets of the WIDER FACE validation dataset, respectively.

\end{abstract}

\section{Introduction}
\label{sec:intro}
Face detection is an essential first step for several computer vision applications like face tracking, face recognition, gender classification and emotion recognition. Its primary objective is the precise localization of face region(s) within an image. Challenges arise particularly in dense crowds (small faces) and adverse conditions such as variations in face pose, low lighting, occlusions, and poor image quality (blur). An optimal face detection system should be able to localize faces in images with high accuracy while operating at low computational costs.

Traditional face detection techniques relied on hand-crafted features along with sliding window techniques~\cite{yow1997feature,wu1999face,TR3}. Among these, the Viola-Jones face detector~\cite{viola2001rapid} have been widely used. Most state-of-art face detection systems are benchmarked on the widely used WIDER FACE dataset~\cite{wider_face2016}. This dataset includes images with various challenging scenarios including blur, pose variations, illumination changes, small faces, and occlusions. Accordingly, the face images are also annotated into easy, medium, and hard categories. Notably, even on the easy subset of the WIDER FACE dataset, the Viola-Jones detector achieves an Average Precision of 41.2\%. This is significantly lesser than the performance of MTCNN (one of the earlier deep network-based proposals), which achieves 85.1\% on the easy subset.

Recent face detection methodologies have leveraged deep learning frameworks for increased precision over traditional methods. These approaches have utilized diverse convolutional neural network (CNN) structures to extract visual features, have incorporated attention modules and improved detection mechanisms. These advancements have yielded substantially improved results on benchmark datasets such as WIDER FACE. Examples of these systems include cascade CNN~\cite{cascadeCNN2015}, RCNN series~\cite{face_rcnn2017}, single-shot face detectors~\cite{s3fd2017,ssh2017}, and RetinaFace~\cite{retinaface}. These face detection systems draw inspiration from the recent advancements in deep learning-based generic object detection methods~\cite{faster_rcnn2015,lin2017focal}. Nevertheless, the performance improvement has led to increased computational demands (FLOPs) for employing these face detectors. This heavy computational requirement arises from utilizing conventional CNN backbones such as ResNet50/101/152~\cite{resnet}, VGG16~\cite{vgg16}, and DenseNet121~\cite{densenet}. Such heavy computation cost makes it hard to deploy such systems for real time applications, especially involving edge devices. Consequently, researchers have focused on the development of lightweight face detection systems. 

\begin{figure}[!htp]
\centering
\includegraphics[width=8cm, height=3.5cm]{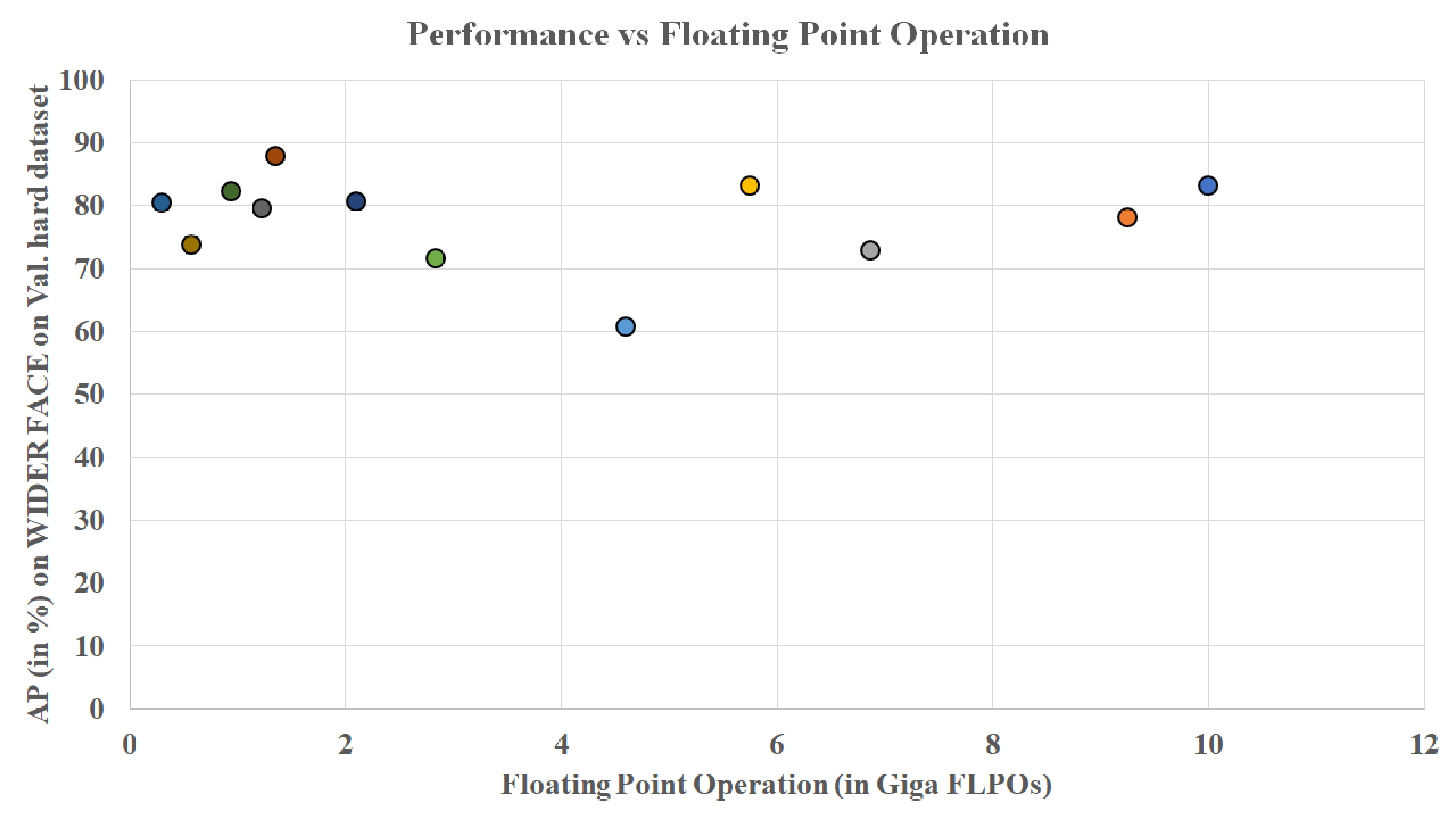}
\includegraphics[width=8cm, height=3.5cm]{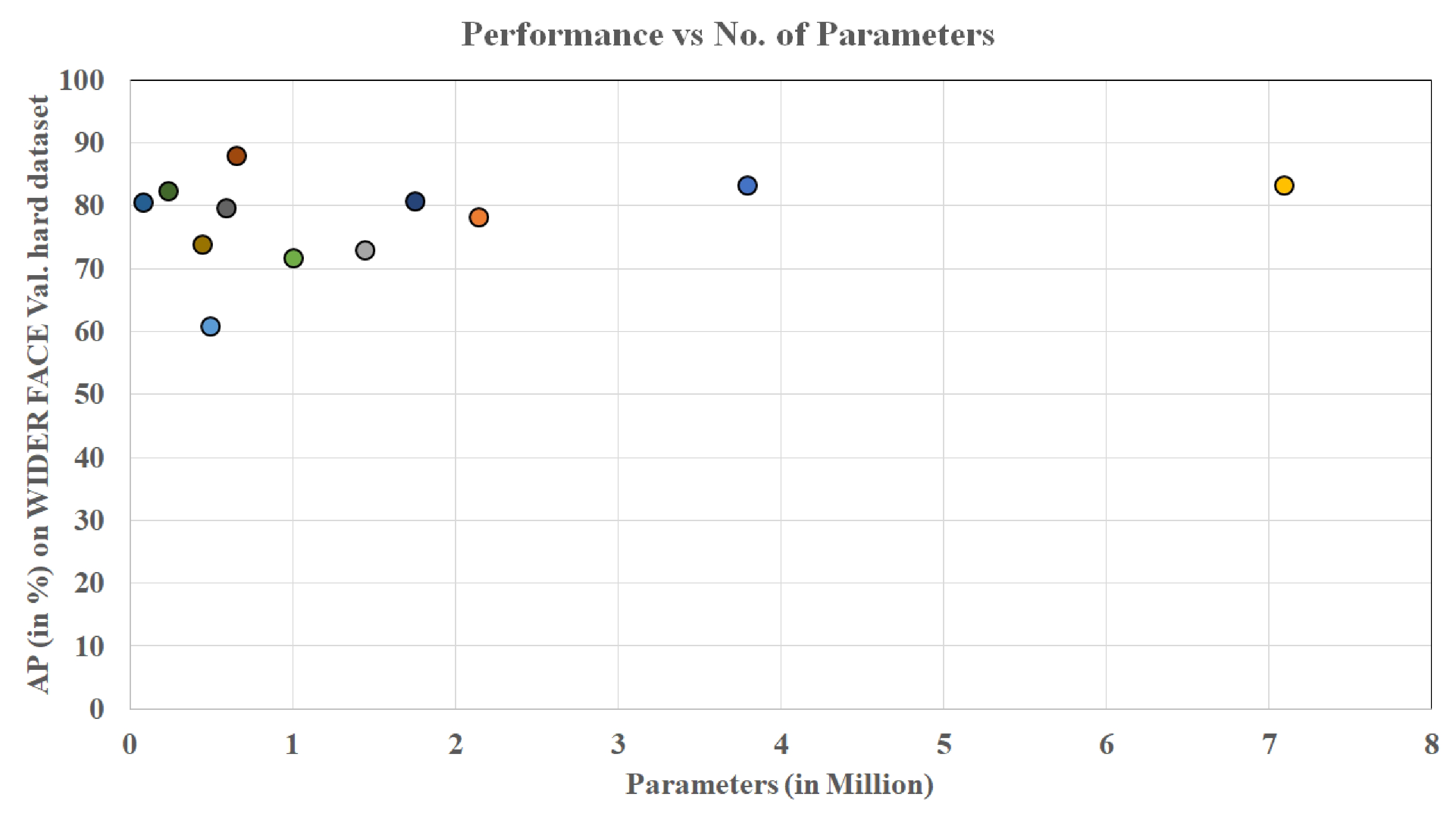}
\includegraphics[width=8cm, height=0.7cm]{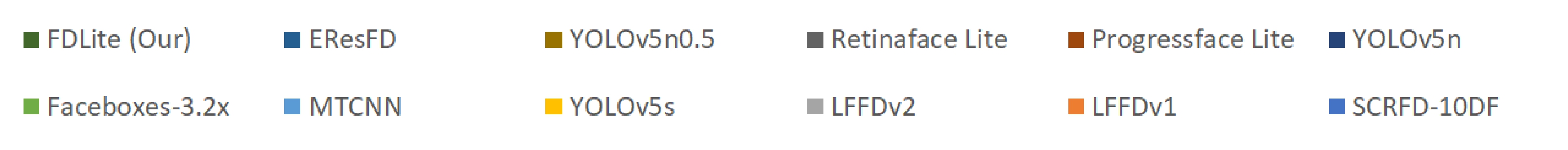}
\caption{\small{Face detection performance (Average Precision) of state-of-art models on \emph{Hard} subset of WIDER FACE validation dataset. The Average Precision is plotted with respect to (a) floating point operations (GFLOPs) and (b) model parameters in millions (M). Note the performance of the proposed face detector -- $82.3\%$ AP with $0.24$M parameters and $0.94$ GFLOPs.}}
\label{fig:into_data}
\end{figure}

Existing works have proposed efficient face detection systems by employing lightweight feature extractor backbones such as the MobileNetV1~\cite{retinaface,progressface} series, ShuffleNetV2~\cite{qi2022yolo5face} series, and others. Additionally, several face detection methods have emerged through design with the help of customized backbones~\cite{2019lffd,Faceboxes,jeong2024EResFD}. These face detection systems have achieved significantly higher accuracy than traditional methods in crowded environments while slightly trailing behind the computation intensive face detectors.

An efficient face detector for real-time applications on edge devices needs to operate with low computation costs without sacrificing accuracy. Accordingly, this work aims to reduce the floating point computations in the network without significantly compromising the face detection accuracy. The proposed face detector FDLite is motivated by the RetinaFace architecture~\cite{retinaface}. It consists of a customized lightweight backbone network (BLite), feature pyramid network (FPN), cascade context prediction modules (CCPM), and detector head (D). Specifically, this work contributes the lightweight customized backbone BLite and the use of two independent multi-task losses. The proposed face detector FDLite is found to provide competitive (or better) performance against 11 state-of-art approaches. The major contributions of this work are as follows: 

\begin{itemize}

\item Proposal of a customized backbone network BLite with $0.167$M parameters and $0.52$ GFLOPs.
\item The use of two independent multi-task losses in the detector head.
\item A lightweight face detector network FDLite with $0.26$M parameters and $0.94$ GFLOPs. It achieves Average Precision (AP) scores of 92.3\%, 89.8\%, and 82.2\% on the easy, medium, and hard subsets of the WIDER FACE validation dataset.

\end{itemize}

\section{Related Work}
\label{sec:formatting}

Several existing deep network based face detectors~\cite{s3fd2017,ssh2017,2018pyramidbox,2019dsfd,srn_face,refineface2020,progressface,liu2020hambox,2022mogface,li2021asfd} are known for high performance but they operate with high computation cost (Table~\ref{tab:related_FD}). Researchers have also proposed several lightweight face detectors with accuracies higher than the classical approaches. This work focuses on the design of lightweight face detectors. Accordingly, only lightweight face detectors are briefly reviewed next.

The cascade CNN based face detectors~\cite{cascadeCNN2015,2016jointCascadeCNN,MTCNN} are considered as lightweight ones due to their low computational requirements. In this framework, candidate windows are initially generated across the input image. A cascade of networks classify these candidate windows as either face or non-face and simultaneously perform bounding box regression while discarding the irrelevant ones. The face prediction is progressively refined through this network cascade. The MTCNN ~\cite{MTCNN} is the most popular among these approaches.

\begin{table*}[!h]
\centering
\caption{Face detection performance (AP in \%) of Computation Intensive and Lightweight face detector networks on the hard subset of the WIDER FACE validation dataset.}
\vspace*{2mm}
\label{tab:related_FD}
\begin{adjustbox}{width=17cm}{
\begin{tabular}{|lcccc|lcccc|}
\hline
\multicolumn{5}{|c|}{\textbf{Computation Intensive Face Detector Networks}} & \multicolumn{5}{c|}{\textbf{Lightweight Face Detector Networks}}                                                         \\ \hline
\multicolumn{1}{|l|}{\textbf{\begin{tabular}[c]{@{}l@{}}Face\\ Detector\end{tabular}}} & \multicolumn{1}{c|}{\textbf{Backbone}} & \multicolumn{1}{c|}{\textbf{AP}} & \multicolumn{1}{c|}{\textbf{\begin{tabular}[c]{@{}c@{}}Param.\\ (in M)\end{tabular}}} & \textbf{GFLOPs} & \multicolumn{1}{l|}{\textbf{\begin{tabular}[c]{@{}l@{}}Face\\ Detector\end{tabular}}} & \multicolumn{1}{c|}{\textbf{Backbone}} & \multicolumn{1}{c|}{\textbf{AP}} & \multicolumn{1}{c|}{\textbf{\begin{tabular}[c]{@{}c@{}}Params.\\ (in M)\end{tabular}}} & \textbf{GFLOPs} \\ \hline
\multicolumn{1}{|l|}{Retinaface~\cite{retinaface}} & \multicolumn{1}{c|}{ResNet152} & \multicolumn{1}{c|}{91.40} & \multicolumn{1}{c|}{80.57} & 249 & \multicolumn{1}{l|}{SCRFD-10DF~\cite{SCRFD}} & \multicolumn{1}{c|}{ResNet18} & \multicolumn{1}{c|}{83.05} & \multicolumn{1}{c|}{3.80} & 10.00 \\ \hline
\multicolumn{1}{|l|}{PyramidBox~\cite{2018pyramidbox}}                                                       & \multicolumn{1}{c|}{VGG16}             & \multicolumn{1}{c|}{88.70}                                                                                       & \multicolumn{1}{c|}{57.18}                                                                  & 236.58                                                                     & \multicolumn{1}{l|}{LFFDv1~\cite{2019lffd}}                                                           & \multicolumn{1}{c|}{Custumized}        & \multicolumn{1}{c|}{78.00}                                                                                       & \multicolumn{1}{c|}{2.15}                                                                   & 9.25                                                                       \\ \hline
\multicolumn{1}{|l|}{RefineFace~\cite{refineface2020}}                                                       & \multicolumn{1}{c|}{ResNet152}         & \multicolumn{1}{c|}{92.00}                                                                                       & \multicolumn{1}{c|}{85.6}                                                                   & 192                                                                        & \multicolumn{1}{l|}{LFFDv2~\cite{2019lffd}}                                                           & \multicolumn{1}{c|}{Custumized}        & \multicolumn{1}{c|}{72.90}                                                                                       & \multicolumn{1}{c|}{1.45}                                                                   & 6.87                                                                       \\ \hline
\multicolumn{1}{|l|}{ASFD~\cite{li2021asfd}}                                                             & \multicolumn{1}{c|}{ResNet101}         & \multicolumn{1}{c|}{92.5}                                                                                        & \multicolumn{1}{c|}{86.1}                                                                   & 183.11                                                                     & \multicolumn{1}{l|}{YOLOv5s~\cite{qi2022yolo5face}}                                                          & \multicolumn{1}{c|}{YOLOv5-CSPNet}     & \multicolumn{1}{c|}{83.10}                                                                                       & \multicolumn{1}{c|}{7.10}                                                                   & 5.75                                                                       \\ \hline
\multicolumn{1}{|l|}{SRN~\cite{srn_face}}                                                              & \multicolumn{1}{c|}{ResNet152}         & \multicolumn{1}{c|}{90.20}                                                                                       & \multicolumn{1}{c|}{81.6}                                                                   & 182                                                                        & \multicolumn{1}{l|}{MTCNN~\cite{MTCNN}}                                                            & \multicolumn{1}{c|}{Custumized}        & \multicolumn{1}{c|}{60.70}                                                                                       & \multicolumn{1}{c|}{0.50}                                                                   & 4.60                                                                       \\ \hline
\multicolumn{1}{|l|}{DSFD~\cite{2019dsfd}}                                                             & \multicolumn{1}{c|}{VGG16}             & \multicolumn{1}{c|}{90.00}                                                                                       & \multicolumn{1}{c|}{141.38}                                                                 & 140.19                                                                     & \multicolumn{1}{l|}{Faceboxes-3.2x~\cite{Faceboxes}}                                                   & \multicolumn{1}{c|}{Custumized}        & \multicolumn{1}{c|}{71.50}                                                                                       & \multicolumn{1}{c|}{1.01}                                                                   & 2.84                                                                       \\ \hline
\multicolumn{1}{|l|}{Progressface~\cite{progressface}}                                                     & \multicolumn{1}{c|}{ResNet152}         & \multicolumn{1}{c|}{91.80}                                                                                       & \multicolumn{1}{c|}{68.63}                                                                  & 123.00                                                                     & \multicolumn{1}{l|}{YOLOv5n~\cite{qi2022yolo5face}}                                                          & \multicolumn{1}{c|}{ShuffleNetV2x1.0}  & \multicolumn{1}{c|}{80.53}                                                                                       & \multicolumn{1}{c|}{1.76}                                                                   & 2.10                                                                       \\ \hline
\multicolumn{1}{|l|}{MOG face~\cite{2022mogface}}                                                         & \multicolumn{1}{c|}{ResNet50}          & \multicolumn{1}{c|}{93.80}                                                                                       & \multicolumn{1}{c|}{34.50}                                                                  & 101.00                                                                     & \multicolumn{1}{l|}{Progressface Lite~\cite{progressface}}                                                & \multicolumn{1}{c|}{MobileNetV1x0.25}  & \multicolumn{1}{c|}{87.90}                                                                                       & \multicolumn{1}{c|}{0.66}                                                                   & 1.35                                                                       \\ \hline
\multicolumn{1}{|l|}{SSH~\cite{ssh2017}}                                                              & \multicolumn{1}{c|}{VGG16}             & \multicolumn{1}{c|}{84.40}                                                                                       & \multicolumn{1}{c|}{19.75}                                                                  & 99.98                                                                      & \multicolumn{1}{l|}{Retinaface Lite~\cite{retinaface}}                                                  & \multicolumn{1}{c|}{MobileNetV1x0.25}  & \multicolumn{1}{c|}{79.50}                                                                                       & \multicolumn{1}{c|}{0.60}                                                                   & 1.23                                                                       \\ \hline
\multicolumn{1}{|l|}{S3FD~\cite{s3fd2017}}                                                             & \multicolumn{1}{c|}{VGG16}             & \multicolumn{1}{c|}{84.00}                                                                                       & \multicolumn{1}{c|}{22.46}                                                                  & 96.6                                                                       & \multicolumn{1}{l|}{YOLOv5n0.5~\cite{qi2022yolo5face}}                                                       & \multicolumn{1}{c|}{ShuffleNetV2x0.5}  & \multicolumn{1}{c|}{73.80}                                                                                       & \multicolumn{1}{c|}{0.45}                                                                   & 0.57                                                                       \\ \hline
\multicolumn{1}{|l|}{HAM BOX~\cite{liu2020hambox}}                                                          & \multicolumn{1}{c|}{ResNet50}          & \multicolumn{1}{c|}{93.30}                                                                                       & \multicolumn{1}{c|}{27.3}                                                                   & 67                                                                         & \multicolumn{1}{l|}{EResFD~\cite{jeong2024EResFD}}                                                           & \multicolumn{1}{c|}{EResNet}           & \multicolumn{1}{c|}{80.41}                                                                                       & \multicolumn{1}{c|}{0.09}                                                                   & 0.30                                                                       \\ \hline
\end{tabular}}
\end{adjustbox}
\end{table*}

The development of single-stage object detection frameworks (such as SSD~\cite{ssd2016} and RetinaNet~\cite{lin2017focal}) led to the proposals of single-stage face detectors~\cite{face_rcnn2017,ssh2017,retinaface} with specific architectural modifications. However, these face detectors utilized computation-intensive backbone networks. Consequently, several lightweight face detection systems have been devised, employing customized backbones like LFFD ~\cite{2019lffd} and FaceBoxes~\cite{Faceboxes} (shown in~\autoref{tab:related_FD}). In FaceBoxes, the incorporation of Rapidly Digested Convolutional Layers (RDCL) facilitated real-time face detection on the CPU, while the integration of Multiple Scale Convolutional Layers (MSCL) allowed for handling faces of various scales by enriching receptive fields. Additionally, a novel anchor densification strategy was introduced to enhance the recall rate of small faces. Meanwhile, LFFD~\cite{2019lffd} introduced a novel customized backbone and presented a receptive field (RF) anchor-free strategy aimed at overcoming the limitations associated with previous anchor-based~\cite{ssd2016,s3fd2017} ones. At that time, these networks~\cite{Faceboxes,2019lffd} achieved the best accuracy in the lightweight face detector category (greater than 70\% AP on the hard subset of the WIDER FACE validation dataset) with less than 10 GFLOPs (shown in~\autoref{tab:related_FD}).

The emergence of classification networks such as MobileNetV1 and V2~\cite{mobilenets} is notable. These networks utilize techniques like depth-wise separable convolution and inverted bottleneck blocks. This development has led to the creation of lighter versions of backbone networks like MobileNetV1x0.25~\cite{mobilenets}. After the introduction of lightweight backbone networks, RetinaFace~\cite{retinaface} and Progressiveface~\cite{progressface} integrated lighter adaptations of MobileNetV1 (MobileNetV1x0.25). These networks~\cite{retinaface,progressface} achieved top accuracy in the lightweight face detection segment, with approximately 88\% AP on the hard set of the WIDER FACE validation dataset, all within a computation of less than 1.5 GFLOPs. A face detector based on YOLOv5 architecture~\cite{qi2022yolo5face} (YOLOv5n0.5) introduced a novel face detection model by employing a lighter variant of the ShuffleNetV2 network (ShuffleNetV2X0.5)~\cite{zhang2018shufflenet}. This network utilized only 0.56 GFLOPs but archived approximately 73\% AP on the hard set of the WIDER FACE validation dataset (shown in~\autoref{tab:related_FD}). Recently, the face detector EResFD~\cite{jeong2024EResFD} achieved the lowest computation cost while maintaining good accuracy (80.43\% AP) on the hard set of the WIDER FACE~\cite{wider_face2016} validation subset, albeit exhibiting degraded accuracy on Easy and Medium subsets (as shown in~\autoref{tab:related_FD}) (less than 90\% AP). Efforts to reduce face detectors persist, but lightweight versions remain critical for edge devices, aiming for lower GFLOPs while maintaining accuracy across various faces of different sizes.

\section{Proposed Work}
\label{sec:proposed}
The proposed face detector FDLite is motivated by the design of RetinaFace~\cite{retinaface}. Accordingly, FDLite has the following key components -- (a) a customized backbone (BLite) network (Subsection~\ref{subsec:new_backbone}) for extracting image features, (b) a Feature Pyramid Network (FPN)~\cite{FPN_network}, (c) Cascade Context Prediction Modules (CCPM)~\cite{jeong2024EResFD}, and (d) the Detector Head (D).

\begin{figure}[htbp]
\centering
\includegraphics[width=0.47\textwidth, height=3cm]{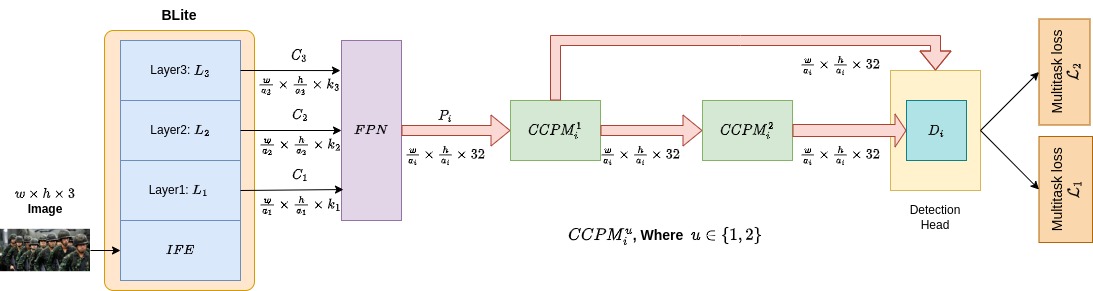}
\caption{\small{Illustrating the key components of the network architecture of the FDLite Face detector. Here, $a^i = 4\times2^i$ where $i \in \{1,2,3\}$}}
\label{fig:Backbone}
\end{figure}

The customized backbone BLite (Subsection~\ref{subsec:new_backbone}) is utilized for spatial feature extraction from input image $\mathbf{I}$ (of size $w\times h\times 3$). BLite is pre-trained with the ImageNet1K dataset~\cite{imagenet}. The Feature Pyramid Network FPN accepts spatial feature maps from intermediate convolutional layers of BLite to provide enhanced feature maps $\mathbf{P}_i$ ($i \in \{1,2,3\}$) of different spatial resolutions. The FPN enriches semantic information by enhancing the edges and corners while bringing out the structural characteristics of face outlines~\cite{FPN_network}. The three feature map outputs of FPN ($\mathbf{P_i}$, $i\in\{1,2,3\}$) are processed through their corresponding Cascade Context Prediction Modules $CCPM^{u}_i$ ($u \in {1,2}$). The first module $CCPM^{1}_i$ receives $\mathbf{P}_i$ as input. The output of $CCPM^{1}_i$ is provided as input to $CCPM^{2}_i$. The CCPM enhances the capability to detect smaller facial features. Subsequently, the refined feature maps obtained from $CCPM^{1}_i$ and $CCPM^{2}_i$ are integrated into the corresponding detector head $D_i$. Each detector head consists of the following three sub-networks for -- (a) face classification task, (b) face bounding box localization, and (c) five facial landmark detection.

\subsection{BLite: The Customized Backbone}
\label{subsec:new_backbone}

A major contribution of this work is the proposal of the customized backbone BLite (shown in Figure~{fig:Backbone}). The input image $\mathbf{I}$ (of size $w \times h \times 3$) is first processed by an Initial Feature Extractor ($IFE$) layer to generate an initial feature tensor $\mathbf{C}_{in} \in \mathbb{R}^{\frac{w}{4}\times \frac{h}{4} \times k_{in}}$. The $IFE$ layer consists of a cascade of one convolutional unit $CBL(m\times n\times k@q;s,p,g)$ and two bottleneck units $CDw(k,q,s)$. Here, $CBL(m\times n\times k@q;s,p,g)$ refers to the application of $q$ number of $m \times n \times k$ convolution kernels with stride $s$, padding $p$, and group convolution parameter $g$ (Notably, $g=1$ signifies no group convolution) followed by batch normalization, and LeakyRelU activation. $CDw(k,q,s)$ consists of two $CBL$ units in cascade -- $CBL(1 \times 1 \times k@q;1,0,1)$ followed by $CBL(3 \times 3 \times q@q;s,1,q)$. Here, $k$ is the channel dimension of the input feature map, and $q$ is that of the output feature map. Note that only the second $CBL$ unit employs a group convolution with $g = q$ with stride $s$.

The initial feature tensor $\mathbf{C}_{in}$ is further refined through three layers $L_{1}, L_{2}$ and $L_{3}$. The feature tensor $\mathbf{C}_{i} \in \mathbb{R}^{ \frac{w}{2^{i+1}} \times \frac{h}{2^{i+1}} \times k_{i} }$ is processed by the layer $L_i$ to produce $\mathbf{C}_{i+1} \in \mathbb{R}^{ \frac{w}{2^{i+2}} \times \frac{h}{2^{i+2}} \times k_i }$. The $CBL$ blocks and Feature Refinement Units ($FRU$) are connected in cascade within each layer $L_i$. The design of $FRU$ is motivated by that of inception module~\cite{Inception_V1_2015} and has residual connections~\cite{resnet} between input and output. The $FRU$ with an input feature tensor of $k$ channels is designated by $FRU(k)$. It does not change the input feature tensor's spatial resolution and channel dimension. The network also uses Max-Pooling units to reduce the spatial resolution of the feature tensors. A $m \times n$ Max-Pooling unit with stride $s$ and padding $p$ is denoted as $MP(m\times n;s,p)$.

The $FRU$ module processes the input feature map using convolution kernels at multiple scales. This allows the network to discern patterns across different resolutions. The resulting features are amalgamated via depth concatenation (refer to Figure~\ref{fig:Backbone}). Initially, the feature map $F_{in}$ undergoes convolution with LeakyReLU activation. This is refered as $CL(3\times 3\times k_{in}@k_{in};1,1,1)$\footnote{$CL(m\times n\times k@q;s,p,g)$ denotes a convolution operation utilizing $q$ number of $m \times n \times k$ kernels with a stride $s$, padding $p$, and a group convolution parameter $g$ (where $g=1$ indicates no group convolution).}. Subsequently, the output of $CL(3\times 3\times k_{in}@k_{in};1,1,1)$ serves as input to two convolutional layers, namely $CL(3\times 3\times k_{in}@\frac{k_{in}}{2};1,1,1)$ and $CL(1\times 1\times k_{in}@\frac{k_{in}}{2};1,0,1)$. The outputs of these layers are then concatenated along the channel dimensions. This concatenated feature map is refined by a $CL(3\times 3\times k_{in}@k_{in};1,1,1)$ convolutional layer. Finally, a residual connection is established by adding the initial feature map $F_{in}$ to the refined feature map, thereby addressing the vanishing gradient issues.

\begin{figure}[htbp]
\centering
\includegraphics[width=0.47\textwidth, height=6cm]{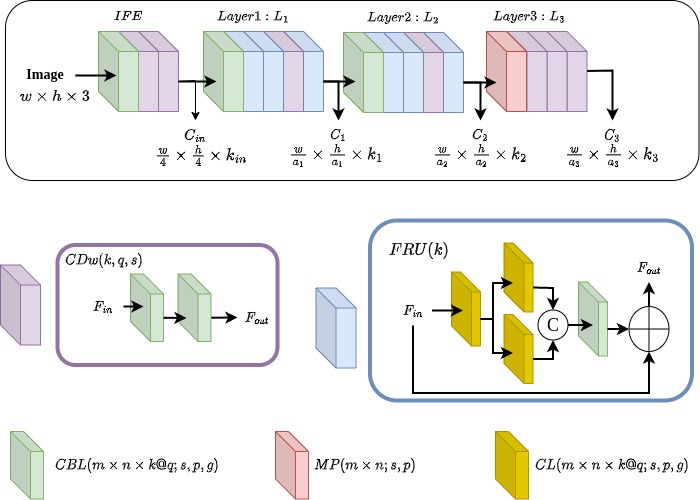}
\caption{\small{Illustrating the architecture of the customized backbone BLite along with its component units ($CBL$, $CDw$, $FRU$, $CL$ and $MP$).}}
\label{fig:Backbone}
\end{figure}

The proposed backbone BLite consisting of the $IFE$ and three layers ($L_1$, $L_2$, $L_3$) (Figure~\ref{fig:Backbone}) is described as follows\footnote{The details of the BLite backbone in terms of number of parameters and floating point operations are presented in Table 1 of the supplementary material.}. \\
\noindent \textbf{Initial Feature Extractor} ($IFE$) -- It has a cascade of one $CBL$ unit $CBL(7\times 7\times 3@8;2,3,1)$ along with two $CDw$ units ($CDw(8,16,1)$ and $CDw(16,32,2)$). The output of $IFE$ ($\mathbf{C}_{in} \in \mathbb{R}^{\frac{w}{4} \times \frac{h}{4} \times 32}$) is fed as input to $L_1$.\\
\noindent \textbf{Layer 1} ($L_1$) -- It has a cascade of one CBL unit ($CBL(3\times 3\times 32@64;2,1,1)$), two $FRU$ units ($2 \times FRU(64)$), one $CDw$ unit ($CDw(64,64,1)$) and another $FRU$ unit ($FRU(64)$). The output of $L_1$ ($C_1 \in \mathbb{R}^{\frac{w}{8} \times \frac{h}{8} \times 64})$ is fed as input to $L_2$ and FPN.\\
\noindent \textbf{Layer 2} ($L_2$) -- It has a cascade of one CBL unit ($CBL(3\times 3\times 64@128;2,1,1)$), two $FRU$ units ($2 \times FRU(128)$), one $CDw$ unit ($CDw(128,128,1)$) and another $FRU$ unit ($FRU(128)$). The output of $L_2$ ($C_2 \in \mathbb{R}^{\frac{w}{16} \times \frac{h}{16} \times 128})$ is fed as input to $L_3$ and FPN.\\
\noindent \textbf{Layer 3} ($L_3$) -- It has a cascade of one max-pooling ($MP(3\times 3;2,1)$) along with three $CDw$ units ($CDw(128,128,1)$, $CDw(128,256,1)$, and $CDw(256,256,1)$). The output of $L_3$ ($C_3 \in \mathbb{R}^{\frac{w}{32} \times \frac{h}{32} \times 256})$ is fed as input to FPN.\\
The feature map $\mathbf{C}_i$ obtained from layer $L_i$ of BLite is fed to the FPN to get an enhanced feature $\mathbf{P}_i$. It is further refined through the CCPM modules ($CCPM_i^{1}$ and $CCPM_i^{2}$) whose output is fed to the detector head $D_i$.

\subsection{Detector Head}
\label{subsec:det_head}

\begin{figure}[htbp]
\centering
\includegraphics[width=0.47\textwidth, height=4cm]{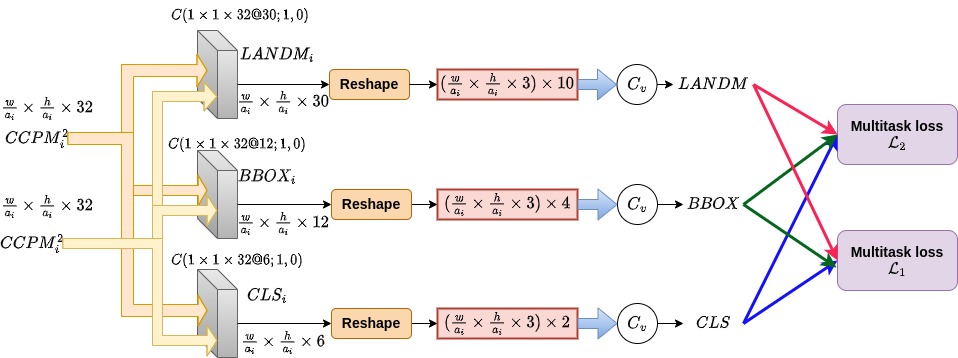}
\caption{\small{The architecture of the detector head includes a sub-network featuring a convolution layer $C(1\times 1\times 32@3\times x;1,0,1)$, where $x$ represents a number of convolution filters (2, 4, and 10) for classification, bounding box regression, and landmark regression, respectively.}}
\label{fig:Detector_head}
\end{figure}

The $i^{\text{th}}$ detector head $D_i$ consists of the following three sub-networks. First, a classification sub-network ($CLS_{i}$) trained with cross-entropy loss to differentiate between faces and non-faces. Second, a sub-network responsible for determining the coordinates of the face bounding boxes. This is known as the bounding-box regression head ($BBOX_{i}$) and is trained using the SmoothL1 loss~\cite{retinaface}. Third, a sub-network dedicated to the localization of five facial landmark coordinates of detected faces. This is named the landmark regression head ($LANDM_{i}$) and is trained by using the SmoothL1 loss. The consolidated output from each task-specific sub-networks ($CLS_{i}$, $BBOX_{i}$ and $LANDM_{i}$) across all detection layers ($D_i$) generates a single tensor after reshaping and vertical concatenation operation ($C_v$). These tensors ($CLS$, $BBOX$, and $LANDM$) are subsequently used for training the network for corresponding task-specific loss function (as shown in~\autoref{fig:Detector_head}).\\

\subsection{Multi-task losses}
\label{subsec:mult_loss}
Building on prior anchor-based detectors~\cite{s3fd2017,retinaface,progressface}, the goal is to optimize the detection objective by concurrently classifying and regressing anchor boxes, along with landmark point regression. This entails minimizing a multi-task loss for each anchor, denoted as j:

\begin{dmath}
\mathcal{L}_{u} = \mathcal{L}_{cls}^{u}(p_j,\hat{p}_j) + \lambda_1{p}_j\mathcal{L}_{box}^{u}(t_j,\hat{t}_j) + \lambda_2{p}_j\mathcal{L}_{landm}^{u}(l_j,\hat{l}_j)
\label{equ:loss}
\end{dmath}

$\mathcal{L}^{u}_{cls}$, $\mathcal{L}^{u}_{box}$, and $\mathcal{L}^{u}_{landm}$ represent the face classification loss (associated with the detector head $CLS$), bounding box regression loss (associated with the detector head $BBOX$), and landmark regression loss (associated with the detector head $LANDM$), respectively.

The classification loss function $\mathcal{L}^{u}_{cls}(p_j, \hat{p}_j)$ compares actual label $p_j$ of the anchor point $j$ and predicted probability $\hat{p}_j$. If the anchor point is a positive example of a face, $p_j$ is set to $1$, and otherwise set to $0$. The binary cross-entropy is used to compute classification loss $\mathcal{L}^{u}_{cls}$. 

The face bounding box regression loss for the $j^{th}$ positive anchor is denoted as $\mathcal{L}^{u}_{box}(t_j, \hat{t}_j)$~\cite{retinaface}. The variables  $t_j=\{t_x, t_y, t_w , t_h\}$ and $\hat{t}_j=\{\hat{t}_x, \hat{t}_y, \hat{t}_w,\hat{t}_h\}$ represent the \{center-abscissa, center-ordinate, width, height\} of the ground-truth bounding box and predicted bounding box respectively. This work uses the bounding box regression loss proposed in~\cite{retinaface}.

The landmark regression loss $\mathcal{L}^{u}_{landm}(l_j, \hat{l}_j)$ is similar to $\mathcal{L}^{u}_{box}$~\cite{retinaface} with five landmark points. Here, $l_j=\{(l^{x1}_j, l^{y1}_j), \ldots  (l^{xm}_j, l^{ym}_j), \ldots  (l^{x5}_j, l^{y5}_j) \}$ and $\hat{l}_j=\{(\hat{l}^{x1}_j, \hat{l}^{y1}_j), \ldots  (\hat{l}^{xm}_j, \hat{l}^{ym}_j), \ldots  (\hat{l}^{x5}_j, \hat{l}^{y5}_j) \}$ are the respective coordinates of ground-truth and predicted facial landmark points. Facial landmark regression employs a target normalization approach based on the anchor center, which is similar to the bounding box center regression. This work uses the landmark regression loss proposed in~\cite{retinaface}.

The FDLite face detector employs the sliding anchor technique~\cite{faster_rcnn2015} for multi-task learning, wherein a predefined set of bounding boxes (referred to as anchor boxes) of various scales are systematically slided across an image. These anchor boxes serve as reference templates to cover faces of different sizes and aspect ratios. Employing the sliding anchor technique enhances the recall rate of face detection.

The proposed detector FDLite utilizes two independent multi-task losses ($\mathcal{L}_{u}$, $u \in \{1,2\}$) to facilitate multi-level~\cite{retinaface} face classification and face localization in an end-to-end framework. The output feature map of $CCPM^{1}_i$ is fed as input to $D_i$ and the resulting tensors are used for computing the multi-task loss $\mathcal{L}_{1}$. Similarly, the output feature map of $CCPM^{2}_i$ is fed as input to $D_i$, and the resulting tensors are used for computing the multi-task loss $\mathcal{L}_{2}$. The combination of these two losses yields a more precise face prediction. Here, the first multi-task loss $\mathcal{L}_{1}$ predicts the bounding boxes using regular anchor selection techniques~\cite{s3fd2017,ssd2016}. The second multi-task loss $\mathcal{L}_{2}$ refines these classification and regression predictions. However, in this study, both multi-task loss functions independently employ regular anchor selection techniques during training. Despite utilizing the same detector head ($D_i$), the input to the detector head differs: for multi-task loss $\mathcal{L}_{1}$, it is sourced from $CCPM^{1}_i$, whereas for multi-task loss $\mathcal{L}_{2}$, it comes from $CCPM^{2}_i$. The proposed framework utilizes multi-task learning for whole network optimization, which integrates several tasks into a unified model. So finally, the combined multi-task losses, $\mathcal{L}_{1}$ and $\mathcal{L}_{2}$, are minimized for any given training anchor $j$ (as elaborated in~\autoref{sec:expSetup}).

\begin{dmath}
\mathcal{L}_{Total} = \mathcal{L}_{1} + \mathcal{L}_{2}
\label{equ:loss_total}
\end{dmath}

\section{Experimental Setup}
\label{sec:expSetup}

\noindent\textbf{Baseline Models --} The performance of FDLite is compared against $11$ state-of-art models. These are RetinaFace-Lite~\cite{retinaface}, Progressiveface~\cite{progressface}, SCRFD-10DF~\cite{2019dsfd}, MTCNN~\cite{MTCNN}, Faceboxes-3.2x~\cite{Faceboxes}, LFFDv1~\cite{2019lffd}, LFFDv2~\cite{SCRFD}, YOLOv5s ~\cite{qi2022yolo5face}, YOLOv5n~\cite{qi2022yolo5face}, YOLOv5n0.5~\cite{qi2022yolo5face} and EResFD~\cite{jeong2024EResFD}. Table~\ref{tab:performance_with_SOA} presents the comparative performance analysis results.\\

\noindent\textbf{Datasets --} FDLite is tested using two standard datasets -- WIDER FACE~\cite{wider_face2016} and FDDB~\cite{fddb2010}. FDLite is trained and validated using the WIDER FACE dataset and FDDB is only used for testing. A multi-scale testing strategy is used to evaluate the results on WIDER FACE~\cite{ssh2017}, whereas the original images are used for the evaluation on FDDB. The WIDER FACE dataset includes 32,203 images with 393,703 annotated bounding boxes outlining faces. These images were randomly sampled from 61 diverse scene categories, presenting various challenges such as pose, scale, occlusion, expression, and illumination variations. The dataset is split into train, validation, and test subsets, comprising 12,883, 3,226, and 16,094 images, respectively. Moreover, five facial landmark points~\cite{retinaface} are utilized during training. Conversely, the FDDB dataset consists of 2,845 images with 5,171 annotated bounding boxes delineating faces, capturing variations in poses and occlusions.\\

\noindent\textbf{Anchor Setting --} At each detection layer ($i \in {1, 2, 3}$), three distinct anchor sizes are employed at every location in the input image. The anchor sizes are determined relative to the original image size as $2^{i}a_i$, $\frac{3}{2}\times2^{i}a_i$, and $2^{i+1}a_i$. Here, $a_i = 4*2^i$ represents the down-sampling factor of each detection layer. These anchors maintain a $1:1$ aspect ratio, covering areas ranging from $16\times16$ to $512\times512$ pixels in the input image. In the training phase, anchors are classified based on their overlap with ground-truth boxes, using the intersection over union (IoU) metric. In the case of multi-task loss $\mathcal{L}_{1}$, anchors surpassing an IoU threshold of $0.7$ are labeled as face anchors, while those falling below $0.3$ are classified as non-face anchors, with other anchors disregarded during training. Conversely, for multi-task loss $\mathcal{L}_{2}$, anchors exceeding a threshold (here set to $0.35$) are designated as face anchors, while the rest are labeled as background or negative. Notably, most anchors (over 99\%) are classified as negative. To mitigate the substantial imbalance between positive and negative examples during training, online hard example mining (OHEM) \cite{OHME} is employed in both multi-task losses. This involves sorting negative anchors based on their loss and selecting the highest-ranking ones. This selection process ensures that the ratio between negative and positive samples is maintained at a minimum of $7:1$ in both multi-task losses~\cite{retinaface}.\\

\noindent\textbf{Training Details --} The proposed face detector is trained using the SGD optimizer, starting with a learning rate of $1\times10^{-3}$, a momentum factor of $0.9$, and a weight decay of $5\times10^{-4}$. The training is conducted over 130 epochs, while the learning rate is reduced by a factor of $10$ at epochs 100 and 120. The training process utilized NVIDIA Tesla V100 GPUs with a batch size of 8.\\

\noindent\textbf{Testing Details --}The performance of FDLite on the WIDER FACE dataset is computed by following standard evaluation procedures~\cite{ssd2016,s3fd2017}. For testing on WIDER FACE, we follow the standard practices of [36, 68] and employ flip as well as multi-scale (the short edge of the image at [500, 800, 1100, 1400, 1700]) strategies. The face confidence scores are acquired for all anchors through the classification sub-networks within the detector head. Subsequently, anchors with confidence scores surpassing the threshold of $0.02$ are chosen for the face detection process. Finally, the non-maximum suppression (NMS) algorithm is applied, using a Jaccard overlap of $0.4$\cite{retinaface}. This algorithm generates the final results by selecting the top $750$ highly confident detections for each image~\cite{refineface2020}.\\

\section{Results and Discussion}
\label{sec:expRes}
This section provides a comprehensive assessment of the proposed face detector FDLite. The effectiveness of FDLite is assessed by comparing its performance with state-of-the-art models using the WIDER FACE and FDDB benchmark datasets. Additionally, an ablation analysis is presented to study the impact of different model components.\\

\begin{table*}[!htpb]
\centering
\caption{\small{Performance comparison of the proposed model against state-of-art face detectors on WIDER FACE validation set. The parameters (in millions) and floating point operations (in GFLOPs for VGA input image) of all models are also compared. Here, AP implies Average precision.}}
\vspace*{2mm}
\label{tab:performance_with_SOA}
\begin{adjustbox}{width=17cm}{
\begin{tabular}{|l|cccc|c|c|}
\hline
\multicolumn{1}{|c|}{\multirow{2}{*}{\textbf{Face Detector}}} & \multicolumn{4}{c|}{\textbf{\begin{tabular}[c]{@{}c@{}}AP (in \%) on\\ WIDER FACE Validation set\end{tabular}}} & \multirow{2}{*}{\textbf{\begin{tabular}[c]{@{}c@{}}Param.\\ (in million)\end{tabular}}} & \multirow{2}{*}{\textbf{\begin{tabular}[c]{@{}c@{}}Computation\\ (in GFLOPs)\end{tabular}}} \\ \cline{2-5}
\multicolumn{1}{|c|}{}                                        & \multicolumn{1}{c|}{\textbf{\begin{tabular}[c]{@{}c@{}}Easy\end{tabular}}} & \multicolumn{1}{c|}{\textbf{\begin{tabular}[c]{@{}c@{}}Medium\end{tabular}}} & \multicolumn{1}{c|}{\textbf{\begin{tabular}[c]{@{}c@{}}Hard\end{tabular}}} & \textbf{\begin{tabular}[c]{@{}c@{}}Overall\\ mAP (\%)\end{tabular}} &                                                                                         &                                                                                             \\ \hline
SCRFD-10DF~\cite{SCRFD}                                                    & \multicolumn{1}{c|}{95.10}                                                           & \multicolumn{1}{c|}{93.90}                                                             & \multicolumn{1}{c|}{85.88}                                                           & 91.62                                                               & 3.80                                                                                    & 10.00                                                                                       \\ \hline
LFFDv1~\cite{2019lffd}                                                        & \multicolumn{1}{c|}{91.00}                                                           & \multicolumn{1}{c|}{88.10}                                                             & \multicolumn{1}{c|}{78.00}                                                           & 85.70                                                               & 2.15                                                                                    & 9.25                                                                                        \\ \hline
LFFDv2~\cite{2019lffd}                                                        & \multicolumn{1}{c|}{83.70}                                                           & \multicolumn{1}{c|}{83.50}                                                             & \multicolumn{1}{c|}{72.90}                                                           & 80.03                                                               & 1.45                                                                                    & 6.87                                                                                        \\ \hline
YOLOv5s~\cite{qi2022yolo5face}                                                       & \multicolumn{1}{c|}{94.30}                                                           & \multicolumn{1}{c|}{92.60}                                                             & \multicolumn{1}{c|}{83.10}                                                           & 90.00                                                               & 7.10                                                                                    & 5.75                                                                                        \\ \hline
MTCNN~\cite{MTCNN}                                                         & \multicolumn{1}{c|}{85.10}                                                           & \multicolumn{1}{c|}{82.00}                                                             & \multicolumn{1}{c|}{60.70}                                                           & 75.93                                                               & 0.50                                                                                    & 4.60                                                                                        \\ \hline
Faceboxes-3.2x~\cite{Faceboxes}                                                & \multicolumn{1}{c|}{79.80}                                                           & \multicolumn{1}{c|}{80.20}                                                             & \multicolumn{1}{c|}{71.50}                                                           & 77.16                                                               & 1.01                                                                                    & 2.84                                                                                        \\ \hline
YOLOv5n~\cite{qi2022yolo5face}                                                       & \multicolumn{1}{c|}{93.60}                                                           & \multicolumn{1}{c|}{91.50}                                                             & \multicolumn{1}{c|}{80.53}                                                           & 88.54                                                               & 1.76                                                                                    & 2.10                                                                                        \\ \hline
Progressface Lite~\cite{progressface}                                             & \multicolumn{1}{c|}{94.90}                                                           & \multicolumn{1}{c|}{93.50}                                                             & \multicolumn{1}{c|}{87.90}                                                           & 92.10                                                               & 0.66                                                                                    & 1.35                                                                                        \\ \hline
Retinaface Lite~\cite{retinaface}                                               & \multicolumn{1}{c|}{92.20}                                                           & \multicolumn{1}{c|}{91.00}                                                             & \multicolumn{1}{c|}{79.50}                                                           & 87.56                                                               & 0.60                                                                                    & 1.23                                                                                        \\ \hline
YOLOv5n0.5~\cite{qi2022yolo5face}                                                    & \multicolumn{1}{c|}{90.70}                                                           & \multicolumn{1}{c|}{88.10}                                                             & \multicolumn{1}{c|}{73.80}                                                           & 84.20                                                               & 0.45                                                                                    & 0.57                                                                                        \\ \hline
EResFD~\cite{jeong2024EResFD}                                                        & \multicolumn{1}{c|}{89.02}                                                           & \multicolumn{1}{c|}{87.96}                                                             & \multicolumn{1}{c|}{80.41}                                                           & 85.79                                                               & 0.09                                                                                    & 0.30                                                                                        \\ \hline
FDLite (our)                                                       & \multicolumn{1}{c|}{$92.30^{5}$}                                                           & \multicolumn{1}{c|}{$89.90^{6}$}                                                             & \multicolumn{1}{c|}{$82.30^{2}$}                                                           & $88.16^{4}$                                                               & $0.24^{2}$                                                                                    & $0.94^{3}$                                                                                        \\ \hline
\end{tabular}}
\end{adjustbox}
\end{table*}

\noindent\textbf{Results on WIDER FACE Dataset --} The performance of the proposed face detector is compared against $11$ baseline algorithms (Section~\ref{sec:expSetup}). The following observations can be made from the results presented in Table~\ref{tab:performance_with_SOA}.

\begin{itemize}

\item FDLite achieves the respective average precision (AP) scores of 92.3\%, 89.9\%, and 82.1\% on Easy, Medium, and Hard subsets of the WIDER FACE validation dataset.

\item FDLite outperforms all baseline face detection frameworks, with the exception of ProgressiveFace~\cite{progressface}) in terms of performance on the hard subset of the WIDER FACE validation dataset while maintaining lower floating point operations (GFLOPs) and network size (parameters in millions).

\item FDLite has lesser floating point operations (in GFLOPs) compared to all baseline face detectors except EResFD and YOLOv5n0.5. However, FDLite outperforms both EResFD and YOLOv5n0.5 in terms of mAP (Table~\ref{tab:performance_with_SOA}).

\item FDLite has lesser parameters compared to all baseline face detectors except for EResFD. Nonetheless, FDLite notably outperforms EResFD and YOLOv5n0.5 in terms of mAP (Table~\ref{tab:performance_with_SOA}).
\end{itemize}

The FDLite face detector achieved competitive (or better) performance (average precision of 92.3\%, 89.9\% and 82.2\% on easy, medium, and hard subsets of the WIDER FACE validation dataset) with only 0.94G FLOPs and 0.24M parameters with respect to the state-of-art models. 

\noindent\textbf{Results on FDDB Dataset --} FDLite undergoes assessment on the FDDB dataset without additional training to showcase its effectiveness across diverse domains. With 1,000 false positives, FDLite achieves a TPR of 97.86\%, a performance comparable to existing methods.

\begin{table}[!htbp]
\centering
\caption{Effect of pre-training of BLite, CCPM module and multi-task losses in the proposed (FDLite) face detector}
\vspace*{2mm}
\label{tab:ablation}
\begin{adjustbox}{width=\columnwidth}{
\begin{tabular}{|c|c|c|ccc|}
\hline
\multirow{2}{*}{\textbf{\begin{tabular}[c]{@{}c@{}}Pre-trained\\ Backbone\end{tabular}}} & \multirow{2}{*}{\textbf{\begin{tabular}[c]{@{}c@{}}Feature Enhancement\\ Module\end{tabular}}} & \multirow{2}{*}{\textbf{\begin{tabular}[c]{@{}c@{}}Number of\\ Multi-Task Losses\end{tabular}}} & \multicolumn{3}{c|}{\textbf{\begin{tabular}[c]{@{}c@{}}AP (in \%) on\\ WIDER FACE Val. set\end{tabular}}} \\ \cline{4-6} 
                                                                                     &                                                                                            &                                                                                                & \multicolumn{1}{c|}{\textbf{Easy}}       & \multicolumn{1}{c|}{\textbf{Medium}}      & \textbf{Hard}      \\ \hline
Yes                                                                                  & \multirow{2}{*}{FPN + SSH}                                                                 & \multirow{2}{*}{Single}                                                                        & \multicolumn{1}{c|}{92.08}               & \multicolumn{1}{c|}{88.98}                & 79.61              \\ \cline{1-1} \cline{4-6} 
No                                                                                   &                                                                                            &                                                                                                & \multicolumn{1}{c|}{91.18}               & \multicolumn{1}{c|}{87.8}                 & 77.47              \\ \hline
Yes                                                                                  & \multirow{2}{*}{FPN + SSH}                                                                 & \multirow{2}{*}{Two}                                                                           & \multicolumn{1}{c|}{92.63}               & \multicolumn{1}{c|}{90.02}                & 81.6               \\ \cline{1-1} \cline{4-6} 
No                                                                                   &                                                                                            &                                                                                                & \multicolumn{1}{c|}{91.64}                  & \multicolumn{1}{c|}{88.73}                   & 79.80                 \\ \hline
Yes                                                                                  & \multirow{2}{*}{FPN + CCPM}                                                                & \multirow{2}{*}{Single}                                                                        & \multicolumn{1}{c|}{92.31}               & \multicolumn{1}{c|}{89.08}                & 80.05              \\ \cline{1-1} \cline{4-6} 
No                                                                                   &                                                                                            &                                                                                                & \multicolumn{1}{c|}{91.3}                  & \multicolumn{1}{c|}{88.21}                   & 77.94                 \\ \hline
Yes                                                                                  & \multirow{2}{*}{FPN + CCPM}                                                                & \multirow{2}{*}{Two}                                                                           & \multicolumn{1}{c|}{92.3}                & \multicolumn{1}{c|}{89.8}                 & 82.4               \\ \cline{1-1} \cline{4-6} 
No                                                                                   &                                                                                            &                                                                                                & \multicolumn{1}{c|}{91.66}               & \multicolumn{1}{c|}{88.9}                 & 80                 \\ \hline
\end{tabular}}
\end{adjustbox}
\end{table}

\noindent\textbf{Ablation Study --} The following ablation analysis experiments are performed to study the effect of different model components. 

\begin{itemize}

\item \textbf{Effect of pre-trained backbone --} Employing the BLite pre-trained backbone (trained on ImageNet1K dataset) with the FDLite face detector resulted in performance improvements across all four versions (Refer to Table \ref{tab:ablation}). Approximately 1\% , 1\% and 2\% respective performance improvements were observed on the easy, medium, and hard subsets of the WIDER FACE validation set.

\item \textbf{Effect of CCPM module --} Ablation experiments showed that substituting SSH~\cite{ssh2017} with CCPM resulted in slight accuracy improvements across the easy and medium subsets of the WIDER FACE validation set. Additionally, there was a 0.5\% increase in accuracy for the hard subset. This trend persisted across configurations using either single or dual multi-task losses.

\item \textbf{Effect of two multi-task loss --} This ablation experiment examines the effect of employing two multi-task losses on FDLite's performance. Integrating them with the SSH module resulted in slight performance improvements. Approximately 0.5\%, 1\% , 2\% respective improvements were noted on the easy, medium and hard subsets of the WIDER FACE validation dataset. However, the improvements were more significant when the two multi-task losses were used with the CCPM module. Notably, around 2\% performance improvement was observed solely on the hard subset of the WIDER FACE validation set, while the performance on other subsets remained unchanged.

\end{itemize}     

\begin{figure*}
\centering     
\subfigure[Tiny faces, blurred faces, and challenging pose orientations]{\includegraphics[width=0.49\textwidth, height=5cm]{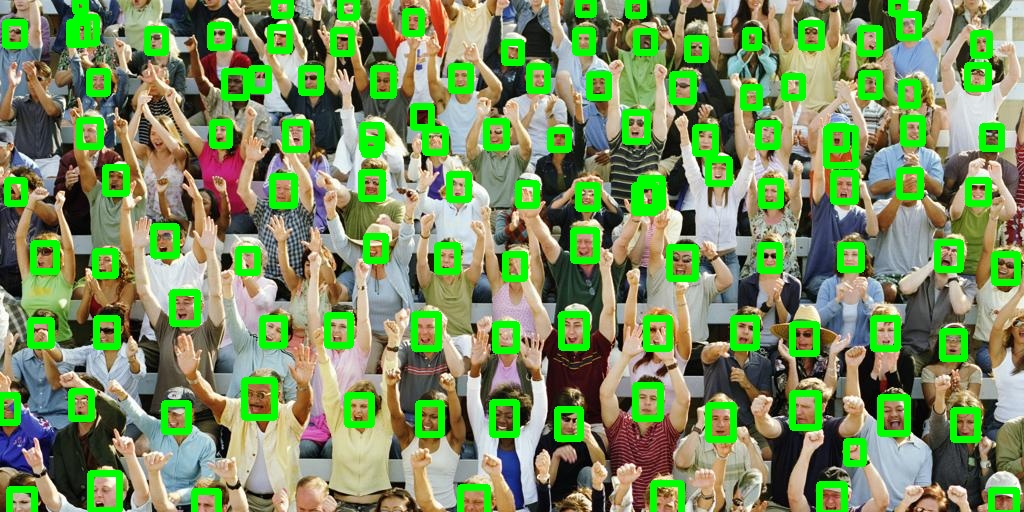}\includegraphics[width=0.49\textwidth, height=5cm]{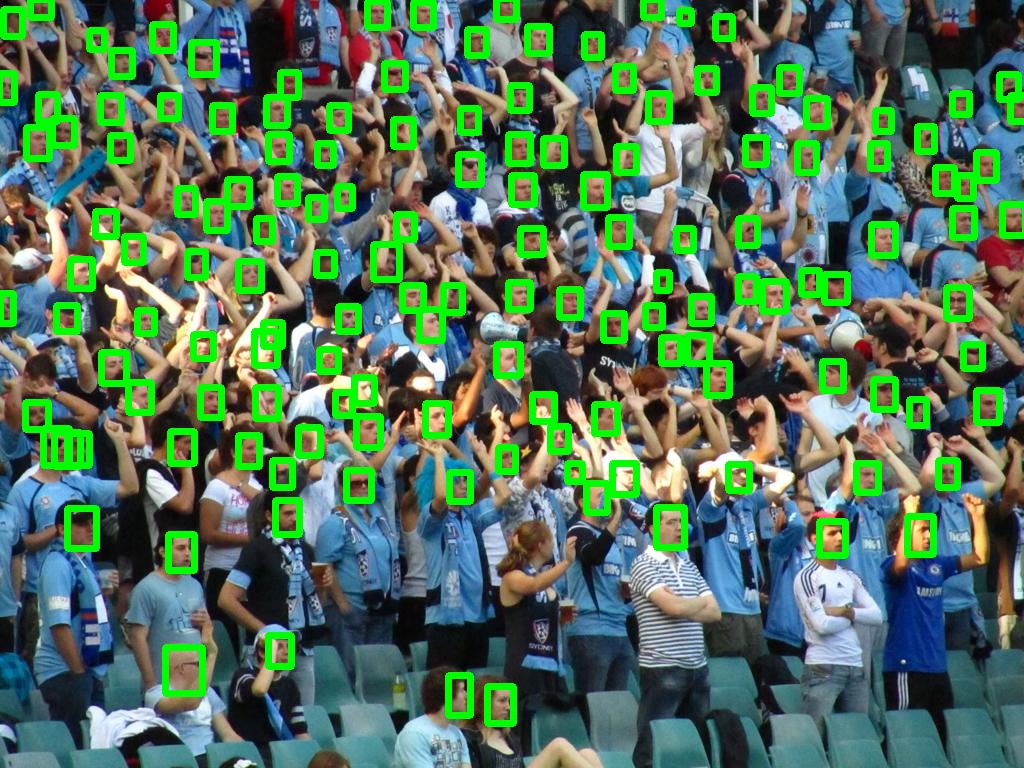}}
\subfigure[Tiny, blurry, occluded faces and challenging pose orientations]{\includegraphics[width=0.49\textwidth, height=5cm]{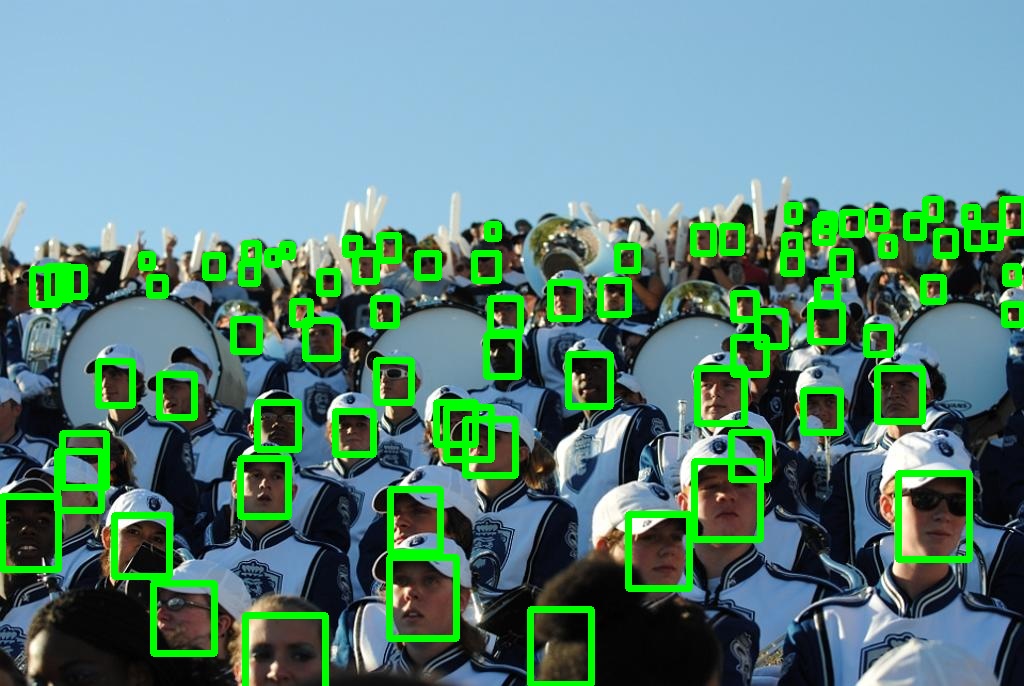}\includegraphics[width=0.49\textwidth, height=5cm]{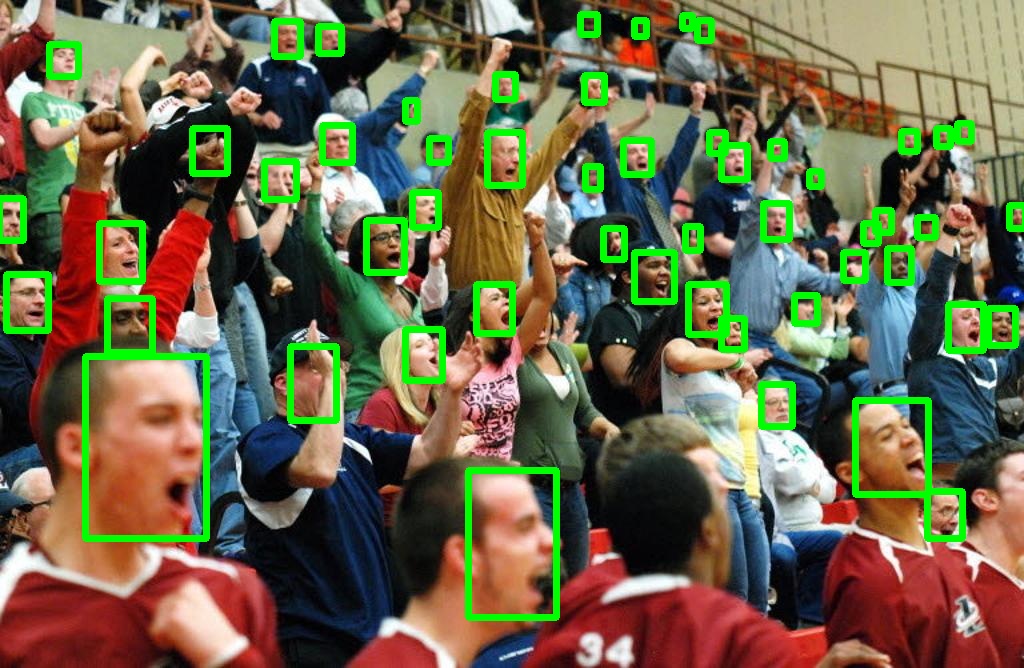}}
\subfigure[Occluded faces]{\includegraphics[width=0.49\textwidth, height=5cm]{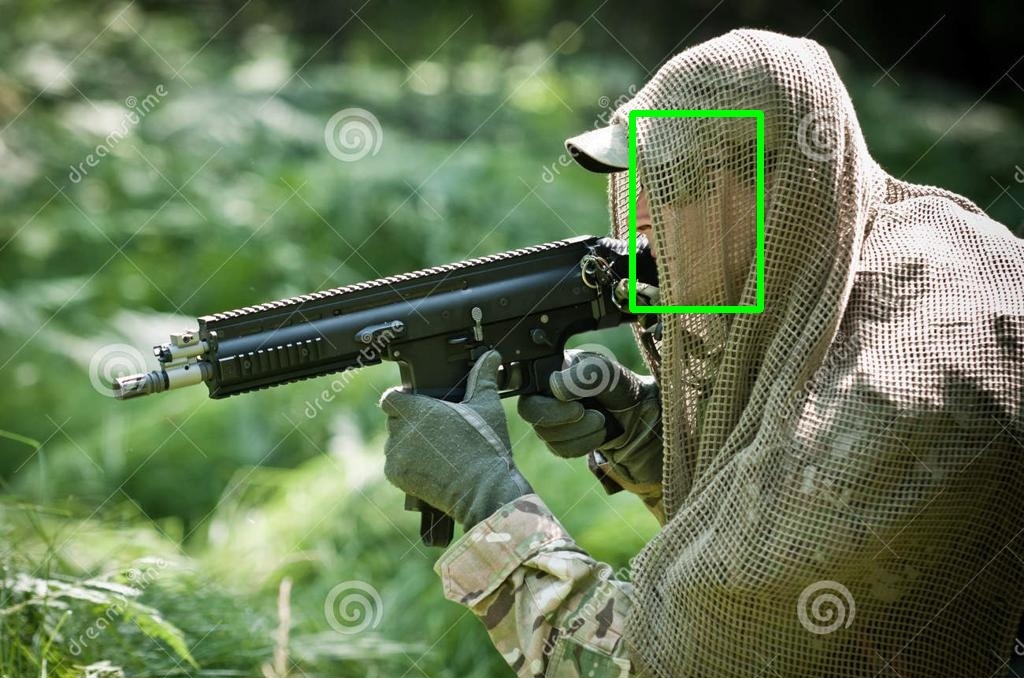}\includegraphics[width=0.49\textwidth, height=5cm]{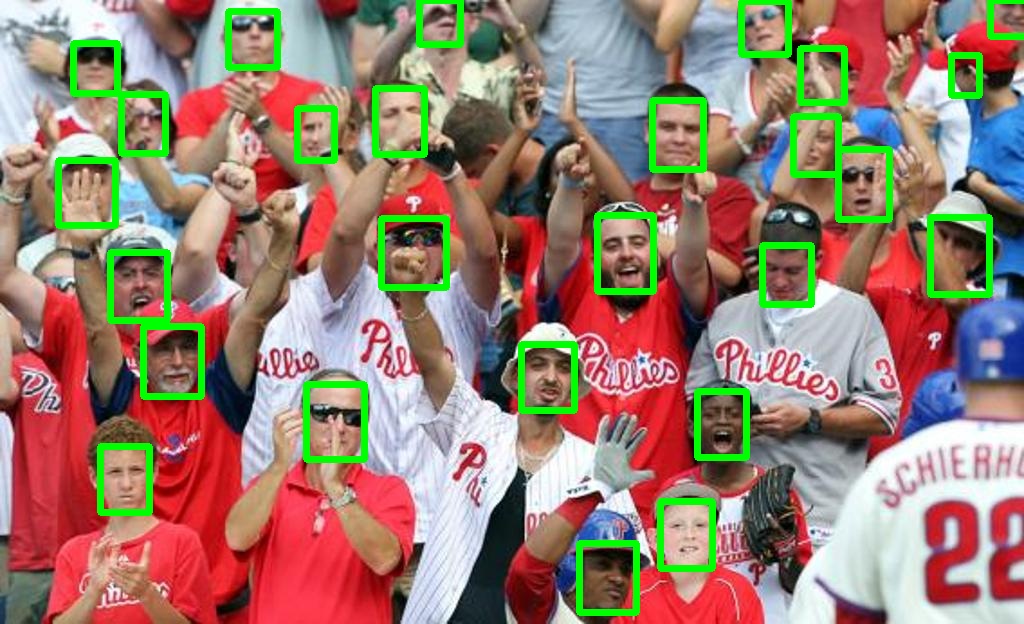}}
\subfigure[Blurred and frontal faces]{\includegraphics[width=0.49\textwidth, height=5cm]{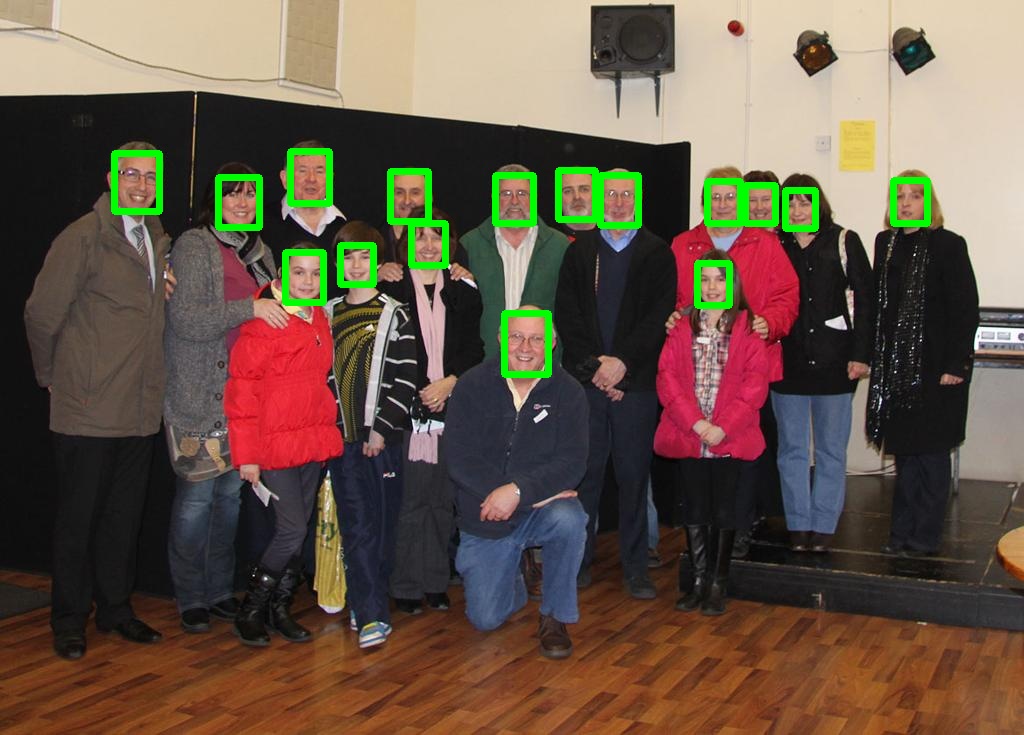}\includegraphics[width=0.49\textwidth, height=5cm]{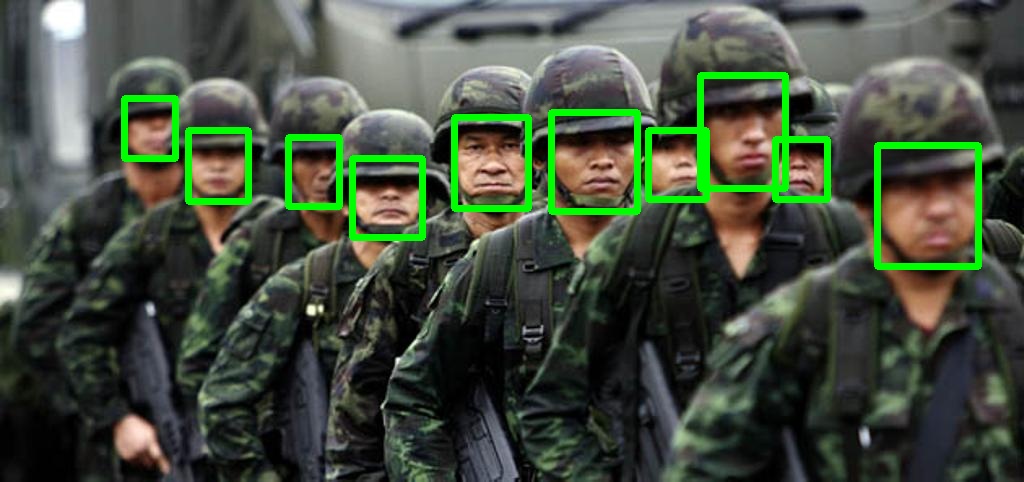}}
\caption{\small{Qualitative results of the proposed face detector's performance under various challenging conditions on WIDER FACE dataset images.}}
\label{fig:reults_on_scenes}
\end{figure*}

\noindent\textbf{Qualitative Performance Analysis --} Figure~\ref{fig:reults_on_scenes} shows the results of face detection in images involving challenging scenarios like occlusions, blur and small faces. These face detection results highlight the effectiveness of the proposed face detector FDLite in overcoming commonly encountered face detection challenges.  

\section{Conclusion}
\label{sec:conc}

This work presented a lightweight face detector FDLite (0.24M parameters and 0.94 GFLOPs) with a novel customized backbone BLite (0.167M parameters and 0.52 GFLOPs). It applied two independent multi-task losses in the face detector heads. The proposal was validated on two standard datasets (WIDER FACE and FDDB) and benchmarked against 11 state-of-the-art approaches. The proposal achieved competitive accuracy with a much lesser number of network parameters and floating point operations. 

This work has focused on reducing the number of network parameters and computations by designing a customized backbone while using standard loss functions and training strategies. Thus, it can be extended by exploring novel loss functions and learning strategies for increasing performance without increasing network complexity. 

{\small
\bibliographystyle{ieee}
\bibliography{egbib}

\begin{thebibliography}{10}\itemsep=-1pt

\bibitem{viola2001rapid}
Paul Viola and Michael Jones.
\newblock Rapid object detection using a boosted cascade of simple features.
\newblock In \emph{Proceedings of the 2001 IEEE Computer Society Conference on Computer Vision and Pattern Recognition. CVPR 2001}, volume~1, pages I--I. IEEE, 2001.

\bibitem{fddb2010}
Vidit Jain and Erik Learned-Miller.
\newblock {FDDB}: A benchmark for face detection in unconstrained settings.
\newblock UMass Amherst technical report, 2010.

\bibitem{wider_face2016}
Shuo Yang, Ping Luo, Chen-Change Loy, and Xiaoou Tang.
\newblock {WIDER FACE}: A face detection benchmark.
\newblock In \emph{Proceedings of the IEEE Computer Society Conference on Computer Vision and Pattern Recognition}, pages 5525--5533, 2016.

\bibitem{cascadeCNN2015}
Haoxiang Li, Zhe Lin, Xiaohui Shen, Jonathan Brandt, and Gang Hua.
\newblock A convolutional neural network cascade for face detection.
\newblock In \emph{Proceedings of the IEEE Computer Society Conference on Computer Vision and Pattern Recognition}, pages 5325--5334, 2015.

\bibitem{2016jointCascadeCNN}
Hongwei Qin, Junjie Yan, Xiu Li, and Xiaolin Hu.
\newblock Joint training of cascaded CNN for face detection.
\newblock In \emph{Proceedings of the IEEE conference on computer vision and pattern recognition}, pages 3456--3465, 2016.

\bibitem{MTCNN}
K. Zhang, Z. Zhang, and Z. Li.
\newblock Joint face detection and alignment using multitask cascaded convolutional networks.
\newblock In \emph{Proceedings of the IEEE signal processing letters}, volume~23, number~10, pages 1499--1503, 2016.

\bibitem{faster_rcnn2015}
Shaoqing Ren, Kaiming He, Ross Girshick, and Jian Sun.
\newblock Faster {R-CNN}: Towards Real-Time Object Detection with Region Proposal Networks.
\newblock \emph{IEEE Transactions on Pattern Analysis \& Machine Intelligence}, volume~39, number~06, pages 1137--1149, 2017.

\bibitem{face_rcnn2017}
Hao Wang, Zhifeng Li, Xing Ji, and Yitong Wang.
\newblock Face {R-CNN}.
\newblock \emph{arXiv preprint arXiv:1706.01061}, 2017.

\bibitem{ssd2016}
Wei Liu et al.
\newblock {SSD}: Single shot multibox detector.
\newblock In \emph{Proceedings of the IEEE European Conference on Computer Vision}, pages 21--37, 2016.

\bibitem{ssh2017}
Mahyar Najibi, Pouya Samangouei, Rama Chellappa, and Larry S. Davis.
\newblock {SSH}: Single stage headless face detector.
\newblock In \emph{Proceedings of the IEEE International Conference on Computer Vision}, pages 4875--4884, 2017.

\bibitem{imagenet}
Jia Deng, Wei Dong, Richard Socher, Li-Jia Li, Kai Li, and Li Fei-Fei.
\newblock ImageNet: A large-scale hierarchical image database.
\newblock In \emph{2009 IEEE Conference on Computer Vision and Pattern Recognition}, pages 248--255, 2009.

\bibitem{s3fd2017}
Shifeng Zhang, Xiangyu Zhu, Zhen Lei, Hailin Shi, Xiaobo Wang, and Stan Z. Li.
\newblock {S3FD}: Single shot scale-invariant face detector.
\newblock In \emph{Proceedings of the IEEE International Conference on Computer Vision}, pages 192--201, 2017.

\bibitem{2018pyramidbox}
Xu Tang, Daniel K. Du, Zeqiang He, and Jingtuo Liu.
\newblock PyramidBox: A context-assisted single shot face detector.
\newblock In \emph{Proceedings of the European Conference on Computer Vision (ECCV)}, pages 797--813, 2018.

\bibitem{2022mogface}
Yang Liu, Fei Wang, Jiankang Deng, Zhipeng Zhou, Baigui Sun, and Hao Li.
\newblock MogFace: Towards a deeper appreciation on face detection.
\newblock In \emph{Proceedings of the IEEE/CVF Conference on Computer Vision and Pattern Recognition}, pages 4093--4102, 2022.

\bibitem{liu2020hambox}
Yang Liu, Xu Tang, Junyu Han, Jingtuo Liu, Dinger Rui, and Xiang Wu.
\newblock Hambox: Delving into mining high-quality anchors on face detection.
\newblock In \emph{2020 IEEE/CVF Conference on Computer Vision and Pattern Recognition (CVPR)}, pages 13043--13051. IEEE, 2020.

\bibitem{li2021asfd}
Jian Li, Bin Zhang, Yabiao Wang, Ying Tai, Zhenyu Zhang, Chengjie Wang, Jilin Li, Xiaoming Huang, and Yili Xia.
\newblock ASFD: Automatic and scalable face detector.
\newblock In \emph{Proceedings of the 29th ACM International Conference on Multimedia}, pages 2139--2147, 2021.

\bibitem{lin2017focal}
Tsung-Yi Lin, Priya Goyal, Ross Girshick, Kaiming He, and Piotr Doll{\'a}r.
\newblock Focal loss for dense object detection.
\newblock In {\em Proceedings of the IEEE International Conference on Computer
  Vision}, pages 2980--2988, 2017.

\bibitem{Faceboxes}
Shifeng Zhang, Xiangyu Zhu, Zhen Lei, Hailin Shi, Xiaobo Wang, and Stan~Z Li.
\newblock Faceboxes: A cpu real-time face detector with high accuracy.
\newblock In {\em 2017 IEEE International Joint Conference on Biometrics
  (IJCB)}, pages 1--9. IEEE, 2017.

\bibitem{progressface}
Jiashu Zhu, Dong Li, Tiantian Han, Lu Tian, and Yi Shan.
\newblock Progressface: Scale-aware progressive learning for face detection.
\newblock In {\em Computer Vision--ECCV 2020: 16th European Conference,
  Glasgow, UK, August 23--28, 2020, Proceedings, Part VI 16}, pages 344--360.
  Springer, 2020.

\bibitem{retinaface}
Jiankang Deng, Jia Guo, Evangelos Ververas, Irene Kotsia, and Stefanos
  Zafeiriou.
\newblock {RetinaFace}: Single-shot multi-level face localisation in the wild.
\newblock In {\em Proceedings of the IEEE/CVF Conference on Computer Vision and
  Pattern Recognition}, pages 5203--5212, 2020.

\bibitem{srn_face}
Cheng Chi, Shifeng Zhang, Junliang Xing, Zhen Lei, Stan~Z Li, and Xudong Zou.
\newblock Selective refinement network for high-performance face detection.
\newblock In {\em Proceedings of the AAAI Conference on Artificial
  Intelligence}, volume~33, number~01, pages 8231--8238, 2019.

\bibitem{refineface2020}
Shifeng Zhang, Cheng Chi, Zhen Lei, and Stan~Z Li.
\newblock {RefineFace}: Refinement neural network for high-performance face
  detection.
\newblock {\em IEEE Transactions on Pattern Analysis \& Machine Intelligence},
  43(11):4008--4020, 2020.

\bibitem{vgg16}
Karen Simonyan and Andrew Zisserman.
\newblock Very deep convolutional networks for large-scale image recognition.
\newblock {\em arXiv preprint arXiv:1409.1556}, 2014.

\bibitem{2019lffd}
Yonghao He, Dezhong Xu, Lifang Wu, Meng Jian, Shiming Xiang, and Chunhong Pan.
\newblock {LFFD}: A light and fast face detector for edge devices.
\newblock {\em arXiv preprint arXiv:1904.10633}, 2019.

\bibitem{2019dsfd}
Jian Li, Yabiao Wang, Changan Wang, Ying Tai, Jianjun Qian, Jian Yang, Chengjie
  Wang, Jilin Li, and Feiyue Huang.
\newblock {DSFD}: dual shot face detector.
\newblock In {\em Proceedings of the IEEE/CVF Conference on Computer Vision and
  Pattern Recognition}, pages 5060--5069, 2019.

\bibitem{Inception_V1_2015}
Christian Szegedy, Wei Liu, Yangqing Jia, Pierre Sermanet, Scott Reed,
  Dragomir Anguelov, Dumitru Erhan, Vincent Vanhoucke, and Andrew Rabinovich.
\newblock Going deeper with convolutions.
\newblock In {\em Proceedings of the IEEE Conference on Computer Vision and
  Pattern Recognition}, pages 1--9, 2015.

\bibitem{SCRFD}
Jia Guo, Jiankang Deng, Alexandros Lattas, and Stefanos Zafeiriou.
\newblock Sample and computation redistribution for efficient face detection.
\newblock {\em arXiv preprint arXiv:2105.04714}, 2021.

\bibitem{resnet}
Kaiming He, Xiangyu Zhang, Shaoqing Ren, and Jian Sun.
\newblock Deep residual learning for image recognition.
\newblock In {\em Proceedings of the IEEE Conference on Computer Vision and
  Pattern Recognition}, pages 770--778, 2016.

\bibitem{densenet}
Gao Huang, Zhuang Liu, Laurens Van Der~Maaten, and Kilian~Q Weinberger.
\newblock Densely connected convolutional networks.
\newblock In {\em Proceedings of the IEEE Conference on Computer Vision and
  Pattern Recognition}, pages 4700--4708, 2017.

\bibitem{mobilenets}
Andrew~G Howard, Menglong Zhu, Bo Chen, Dmitry Kalenichenko, Weijun Wang,
  Tobias Weyand, Marco Andreetto, and Hartwig Adam.
\newblock {MobileNets}: Efficient convolutional neural networks for mobile
  vision applications.
\newblock {\em arXiv preprint arXiv:1704.04861}, 2017.

\bibitem{FPN_network}
Tsung-Yi Lin, Piotr Doll{\'a}r, Ross Girshick, Kaiming He, Bharath Hariharan,
  and Serge Belongie.
\newblock Feature pyramid networks for object detection.
\newblock In {\em Proceedings of the IEEE Conference on Computer Vision and
  Pattern Recognition}, pages 2117--2125, 2017.

\bibitem{qi2022yolo5face}
Delong Qi, Weijun Tan, Qi Yao, and Jingfeng Liu.
\newblock YOLO5Face: why reinventing a face detector.
\newblock In {\em European Conference on Computer Vision}, pages 228--244.
  Springer, 2022.

\bibitem{zhang2018shufflenet}
Xiangyu Zhang, Xinyu Zhou, Mengxiao Lin, and Jian Sun.
\newblock Shufflenet: An extremely efficient convolutional neural network for
  mobile devices.
\newblock In {\em Proceedings of the IEEE conference on computer vision and
  pattern recognition}, pages 6848--6856, 2018.

\bibitem{OHME}
Abhinav Shrivastava, Abhinav Gupta, and Ross Girshick.
\newblock Training region-based object detectors with online hard example
  mining.
\newblock In {\em Proceedings of the IEEE Conference on Computer Vision and
  Pattern Recognition}, pages 761--769, 2016.

\bibitem{wu1999face}
Haiyuan Wu, Qian Chen, and Masahiko Yachida.
\newblock Face detection from color images using a fuzzy pattern matching
  method.
\newblock {\em IEEE Transactions on Pattern Analysis \& Machine Intelligence},
  21(6):557--563, 1999.

\bibitem{yow1997feature}
Kin~Choong Yow and Roberto Cipolla.
\newblock Feature-based human face detection.
\newblock {\em Image and vision computing}, 15(9):713--735, 1997.

\bibitem{TR3}
Rein-Lien Hsu, Mohamed Abdel-Mottaleb, and Anil~K Jain.
\newblock Face detection in color images.
\newblock {\em IEEE transactions on pattern analysis and machine
  intelligence}, 24(5):696--706, 2002.

\bibitem{jeong2024EResFD}
Joonhyun Jeong, Beomyoung Kim, Joonsang Yu, and Youngjoon Yoo.
\newblock EResFD: Rediscovery of the effectiveness of standard convolution for
  lightweight face detection.
\newblock In {\em Proceedings of the IEEE/CVF Winter Conference on Applications
  of Computer Vision}, pages 988--998, 2024.


\end{thebibliography}
}

\end{document}